# CRU: A Novel Neural Architecture for Improving the Predictive Performance of Time-Series Data

Sunghyun Sim, *Member, IEEE, Dohee Kim,* Hyerim Bae, *Member, IEEE*

**Abstract**—The time-series forecasting (TSF) problem is a traditional problem in the field of artificial intelligence. Models such as Recurrent Neural Network (RNN), Long Short Term Memory (LSTM), and GRU (Gate Recurrent Units) have contributed to improving the predictive accuracy of TSF. Furthermore, model structures have been proposed to combine time-series decomposition methods, such as seasonal-trend decomposition using Loess (STL) to ensure improved predictive accuracy. However, because this approach is learned in an independent model for each component, it cannot learn the relationships between time-series components. In this study, we propose a new neural architecture called a correlation recurrent unit (CRU) that can perform time series decomposition within a neural cell and learn correlations (autocorrelation and correlation) between each decomposition component. The proposed neural architecture was evaluated through comparative experiments with previous studies using five univariate time-series datasets and four multivariate time-series data. The results showed that long- and short-term predictive performance was improved by more than 10%. The experimental results show that the proposed CRU is an excellent method for TSF problems compared to other neural architectures.

**Index Terms**—Time-series Data, Time-series forecasting, STL Cell, Correlation Gate, Correlation Recurrent Unit

———————————— ◆ ————————————

## 1 INTRODUCTION

Time-series data refer to records of continuous phenomena of temporal events subdivided into the past, present, and future [1]. These data are constantly generated and stored in different areas, such as medical [2], human activity [3], [4], financial markets [5], [6], and urban traffic control [7], [8]. Studies using time-series data are used for time-series prediction, outlier and trajectory detection, and human pattern analysis, among which the time-series forecasting (TSF) problem is a traditional challenge [9]-[11].

To solve the TSF problem, it is important to accurately predict the value of the prediction point [12]. Traditionally, statistical approaches, such as ARIMA, have been mainly used in TSF problems [13]. In the deep learning domain, time-series prediction models based on sequence models, such as RNN [14], LSTM [15], and GRU [16] have been proposed and they showed better forecasting performance than conventional statistical methods [17]-[19]. Despite these efforts, there is room for improvement in short- and long-term TSF [20]-[22].

---

- *Sunghyun Sim is with theIndustrial Management & Big Data Engineering Major, Division of Industrial Convergence System Engineering, Dong eui University, 176 Eomgwang No, Gaya Dong 24, Busanjin Gu, 47340 Busan, South Korea; Email: ssh@deu.ac.kr*

- *Doohee Kim and Hyerim Bae are majoring in Industrial Data Science & Engineering, Department of Industrial Engineering, Pusan National University, Pusan National University, 30-Jan-jeon Dong, Geum-Jeong Gu, 609-753, Busan, South Korea; Email: kimdohee@pusan.ac.kr; hrbae@pusan.ac.kr (Corresponding author).*

- The CRU implementation code is available at: https://github.com/simsimhyun/CRU

Recently, a method of increasing prediction accuracy by applying a decomposition method such as seasonal-trend decomposition using Loess (STL) has been proposed [23]. Fig. 1 shows that this approah generally divides time-series data into three components (trend, seasonal, and remainder components) and learns a time-series prediction model using those components. The figure shows the general framework of deep learning models used in various studies [25]–[27]. This approach improved prediction accuracy compared to the conventional RNN, LSTM, and GRU approaches. However, it has two limitations. 1) The individual models must be learned depending on the number of decomposed components and the trained models can become heavy. 2) The influence within each decomposition component cannot be learned because each decomposition component is learned in an individual neural network rather than a single cell.

To overcome the limitations of existing studies and increase the prediction accuracy of TSF, this study proposes a new neural structure called a correlation recurrent unit (CRU) that combines a new concept (STL cell) and two new gates (correlation and autocorrelation gates).

The novel neural architecture proposed in this study has the following advantages. First, the input data is divided into time-series decomposition components through the STL cell and the hidden state is updated by considering both autocorrelation and correlation occurring between time-series components, making it possible to predict more accurately. Second, unlike previous studies that used to learn decomposed time-series decomposition components using multiple neural networks, learning is performed in a single cell; therefore, a lighter neural network can be provided to users. The contributions of this study are as follows.:

- (C1) We developed a novel STL cell structure that can perform STL decomposition within a single cell.
- (C2) We developed two new gates (autocorrelation and correlation gates) that can learn autocorrelation within time-series components and the relationship among different time-series components.
- (C3) We combined them to develop a novel neural architecture called CRU, which is suitable for TSF problems.
- (C4) We validated the superiority of the proposed method over the existing methods by improving its forecasting through various experiments using univariate time series data and multivariate time series data.

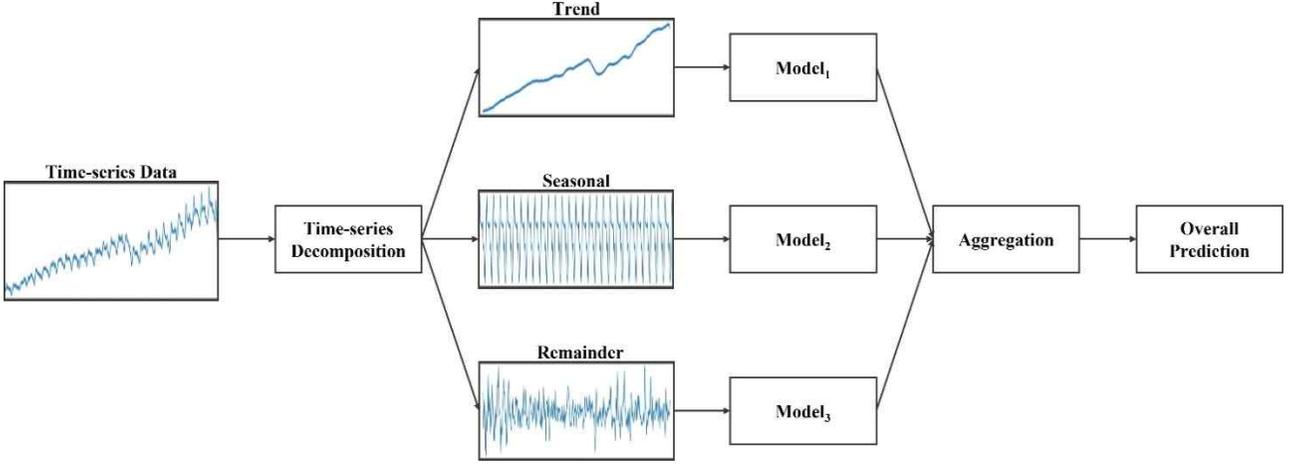

Fig. 1. Example of a deep learning model framework based on time-series decomposition, used in previous studies.

The remainder of this paper is organized as follows. Section 2 presents the study background. Section 3 presents the description of the proposed methodology. Section 4 presents the verification of the proposed neural architecture. Section 5 summarizes the study and discusses future research and applications.

## 2 BACKGROUND

### 2.1 RNN, LSTM, and GRU

Recurrent neural networks (RNNs), long short-term memory (LSTM), and gated recurrent unit (GRU) are approaches to solve TSF problems in deep learning areas [28]. An RNN is a natural generalization of the feedforward neural network to a sequence [29]. The RNN architecture is given by Eq (1).

$$h_t = f(W_{xh}x_t + W_{hh}h_{t-1} + b_h)$$
$$y_t = g(W_{hy}h_t + b_y)$$

(1)

where $W_{xh}$, represents the weight parameter between the input and hidden layers, $W_{hh}$ represents the weight parameters between the input and hidden layers. $W_{hy}$ represents the weight parameters between the hidden and output layers. $b_h$ and $b_y$ represent the bias of the hidden and output layers, respectively. Functions $f(\cdot)$ and $g(\cdot)$ are the activation functions for the hidden and output layers, respectively. Although RNN has several advantages, it has a vanishing gradient problem, which hinders the learning of long sequence data [30].

Long short-term memory (LSTM) was proposed to address the long-term dependencies that RNN cannot handle [31]. With LSTMs, the time step $t$, consists of three gates (input $i_t$, forget $f_t$, and output gates $o_t$) and one memory cell $c_t$. LSTM updates each gate and memory cell using the following equations.

$$i_t = \sigma(W_{xi}x_t + W_{hi}h_{t-1} + b_i)$$
$$f_t = \sigma(W_{xf}x_t + W_{hf}h_{t-1} + b_f)$$
$$o_t = \sigma(W_{xo}x_t + W_{ho}h_{t-1} + b_o)$$
$$\tilde{c}_t = \tanh(W_{xc}x_t + W_{hc}h_{t-1} + b_c)$$
$$c_t = f_t \odot c_{t-1} + i_t \odot \tilde{c}_t$$
$$h_t = o_t \odot \tanh(c_t)$$

(2)

where $W_{xi}, W_{xf}$, and $W_{xo}$ represent the weight parameters between the input layer and each gate, respectively. $W_{xc}$ represents the weight parameter between the input layer and memory cell. $W_{hi}, W_{hf}$, and $W_{ho}$ represent the weight parameters between the hidden layer and each gate, respectively, and $W_{hc}$ represents the weight parameter between the hidden layer and memory cell. $b_i, b_f, b_o$ and $b_c$ represent the bias of each gate and memory cell, respectively. $\sigma(\cdot)$ is a sigmoid activation function. $\odot$ denotes the element-wise product.

Cho et al. proposed the gated recurrent unit (GRU) as an alternative to LSTM; it is similar to LSTM but has a simpler cell architecture [32]. Unlike LSTM, in GRU, a single gate unit controls both forgetting factors and decisions to update the unit state. The GRU updates each gate using the following equation.

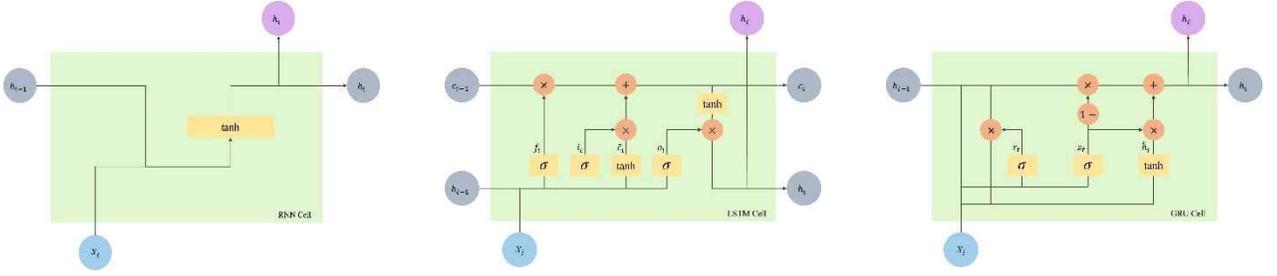

Fig. 2. The Structure of RNN, LSTM, and GRU Cells

$$r_t = \sigma(W_{xr}x_t + W_{hr}h_{t-1} + b_r)$$

$$u_t = \sigma(W_{xu}x_t + W_{hu}h_{t-1} + b_u)$$

$$\tilde{h}_t = \tanh(W_{xh}x_t + W_{hh}(r_t \odot h_{t-1}) + b_h)$$

$$h_t = u_t \odot \tilde{h}_t + (1 - u_t) \odot h_{t-1}$$

(3)

where $r_t$ and $u_t$ denote the reset and update gates, respectively. The structures of the RNN, LSTM, and GRU, are illustrated in Fig. 2.

## 2.2 Seasonal-Trend Decomposition using LOESS

STL is a time-series decomposition method based on loess regression. Loess regression is a nonparametric technique that uses locally weighted regression to fit a smooth curve through points in a scatter plot [33], [34].

STL consists of two procedures [35]: an inner loop and an outer loop. The inner loop is nested inside the outer loop. The main steps of an inner loop are seasonal and trend smoothings. The detailed six steps of the k$^{th}$ epoch are as follows [24].

- Step 1: **Detrending.** Obtain a new series by subtracting the trend values $T_t^k$ from original values $Y_t$.
- Step 2: **Cycle-subseries smoothing.** Each cycle subseries obtained from Step 1 is regressed by loess and the result is recorded as $C_t^{k+1}$.
- Step 3: **Low-pass filtering.** The filter for $C_t^{k+1}$ includes three steps. The first step is the moving average of length $n$, where $n$ is the number of samples. The following step is also a moving average of length $n$. The last step is a moving average of length 3. Subsequently, the loess is applied to the results of low-pass filtering. The result is recorded as $L_t^{k+1}$.
- Step 4: **Detrending the smoothed cycle subseries.** The seasonal series $S_t^{k+1}$ is obtained using Eq. (4)

$$S_t^{k+1} = C_t^{k+1} - L_t^{k+1}, \text{ for } t = 1 \text{ to } n$$

(4)

- Step 5: **Deseasonalizing**. Obtain deseasonalized series using $Y_t - S_t^{k+1}$.

- Step 6: **Trend smoothing.** The trend series $T_t^{k+1}$ is obtained after the loess is applied to the deseasonalized series.

The steps of the outer loop are as follows. The values of $T_t$ and $S_t$ are obtained after the inner loop. Subsequently, the remaining series $R_t$ is calculated according to Eq. (5)

$$R_t = Y_t - T_t - S_t$$

(5)

The robustness weight $\rho$ is defined to evaluate the robustness of $R_t$. $\rho_t$ is the robustness weight at time $t$.

$$\rho_t = \frac{B(|R_t|)}{6 \, median \, (|R_t|)}$$

(6)

The formula for bisquare weight function $B$ is as follows:

$$B(u) = \begin{cases} (1 - u^2)^2 & for \ 0 \le u < 1 \\ 0 & for \quad u > 1 \end{cases}$$

(7)

# 3 PROPOSED NEURAL ARCHITECTURE WITH STL CELL AND AUTOCORRELATION AND CRRELATION GATES

This section introduces the proposed STL cell and describes the structure of combining STL cells with RNN, LSTM, and GRU. Subsequently, we introduce the autocorrelation and correlation gates and propose a new type of neural network, CRU, with the two gates.

## 3.1 Basic Concept of STL cell with RNN

Generally, neural networks which deal with time series have cell structures. In this study, STL cells are placed at the entrance of cells to decompose the time-series data. Using STL cells, time-series data are decomposed into three components: trend, seasonal, and remainder components.

Assuming that the time-series input $X = (x_1, x_2, \ldots, x_T)$, $1 \le t \le T$ is obtained. Unlike the RNN structure, which updates the weight parameters and bias of hidden layers directly using $x_t$, in the proposed method, the $x_t$ first enters the STL cell before updating the hidden state. The STL cell works using the same procedure as that proposed by Cleveland et al. [23]. Through the STL cell, the time-series input $x_t$ is divided into $x_t^t$, $x_t^s$, and $x_t^r$, which are components of trend, seasonality, and remainder, respectively, as shown in Eq. (8).

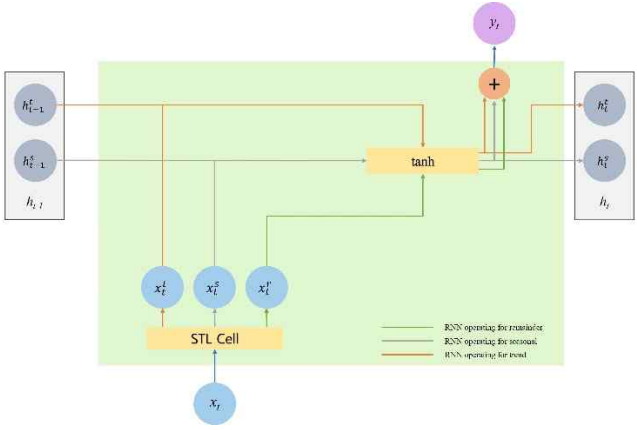

Fig. 3. RNN Structure with STL Cell

$$x_t = x_t^s + x_t^t + x_t^r \tag{8}$$

For the three decomposed inputs, we use three hidden states—$h_t^s$, $h_t^t$ and $h_t^r$—for handling trend, seasonality, and remainder component, respectively. The three hidden states are updated as shown in Eq. (9).

$$
\begin{aligned}
h_t^s &= f(W_{x^s h^s} x_t^s + W_{h^s h^s} h_{t-1}^s + b_{h^s}^t) \\
h_t^t &= f(W_{x^t h^t} x_t^t + W_{h^t h^t} h_{t-1}^t + b_{h^t}^t) \\
h_t^r &= f(W_{x^r h^r} x_t^r + b_{h^r}^r)
\end{aligned}
\tag{9}
$$

The $h_t^t$ and $h_t^s$ (at time $t$) are updated using the $h_{t-1}^t$ and $h_{t-1}^s$ (at time $t$-1). Conversely, the $h_t^r$ can be updated without using $h_{t-1}^r$. This is because the remaining components are not affected by the value at the previous time point, whereas trend and seasonal components are generally affected by the values at the previous time point [36].

The predicted value at time $t$, $\hat{y}_t$, is predicted using the three hidden states ($h_t^s$, $h_t^t$ and $h_t^r$), as shown in Eq. (10).

$$\hat{y}_t = g(W_{h^s y} h_t^s + W_{h^t y} h_t^t + W_{h^r y} h_t^r + b_y) \tag{10}$$

Fig. 3 shows the overall structure of the RNN with STL

cell (RNN-STLC).

## 3.2 LSTM-STLC and GRU-STLC

This subsection introduces new cell structures, LSTM-STLC and STL-GRU, in which STL cells are combined with LSTM and GRU, respectively. LSTM-STLC and GRU-STLC use the existing input, forget, and output gates (LSTM) and reset and update gates(GRU). However, LSTM-STLC and GRU-STLC are different because each hidden state has three corresponding components: trend, seasonality, and remainder.

Similar to RNN-STLC, hidden states of trend and seasonality in LSTM-STLC are updated using the previous hidden states and current input. However, the hidden states of the remainder are updated using only the current input. Thus, the input, forget, and output gates in the LSTM-STLC operate as shown in Eq. (11).

$$
\begin{aligned}
i_t^s &= \sigma(W_{x^s i^s} x_t^s + W_{h^s i^s} h_{t-1}^s + b_{i^s}^s) \\
i_t^t &= \sigma(W_{x^t i^t} x_t^t + W_{h^t i^t} h_{t-1}^t + b_{i^t}^t) \\
f_t^s &= \sigma(W_{x^s f^s} x_t^s + W_{h^s f^s} h_{t-1}^s + b_{f^s}^s) \\
f_t^t &= \sigma(W_{x^t f^t} x_t^t + W_{h^t f^t} h_{t-1}^t + b_{f^t}^t) \\
o_t^s &= \sigma(W_{x^s o^s} x_t^s + W_{h^s o^s} h_{t-1}^s + b_{o^s}^s) \\
o_t^t &= \sigma(W_{x^t o^t} x_t^t + W_{h^t o^t} h_{t-1}^t + b_{o^t}^t)
\end{aligned}
\tag{11}
$$

Each gate value operated in Eq. (11) is used to update the memory cells using trend and seasonal components, as shown in Eq. (12).

$$
\begin{aligned}
\tilde{c}_t^s &= \tanh(W_{x^s c^s} x_t^s + W_{h^s c^s} h_{t-1}^s + b_{c^s}^s) \\
\tilde{c}_t^t &= \tanh(W_{x^t c^t} x_t^t + W_{h^t c^t} h_{t-1}^t + b_{c^t}^t) \\
c_t^s &= f_t^s \odot c_{t-1}^s + i_t^s \odot \tilde{c}_t^s \\
c_t^t &= f_t^t \odot c_{t-1}^t + i_t^t \odot \tilde{c}_t^t
\end{aligned}
\tag{12}
$$

Finally, Eq. (13) updates each hidden state corresponding to the time-series component.

$$
\begin{aligned}
h_t^s &= o_t^s \odot \tanh(c_t^s) \\
h_t^t &= o_t^t \odot \tanh(c_t^t)
\end{aligned}
\tag{13}
$$

GRU-STLC also updates the hidden state using a procedure similar to that of LSTM-STLC, as shown in Eq. (14).

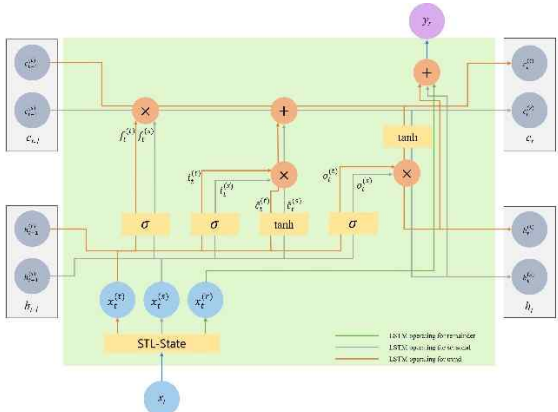

Fig. 4. STL-LSTM Cell Structure

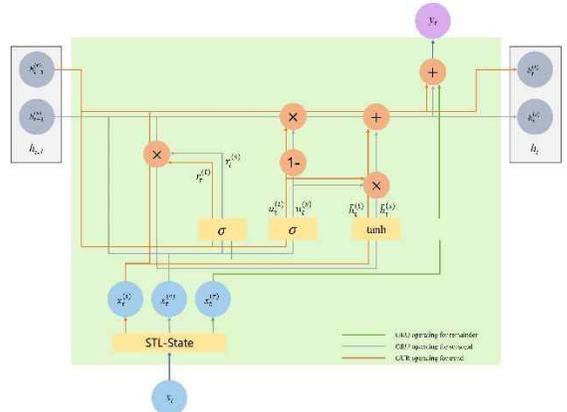

Fig. 5. STL-GRU Cell Structure

The cell structures of LSTM-STLC and GRU-STLC are shown in Fig. 4 and 5, respectively.

$$r_t^s = \sigma(W_{x^s r^s} x_t^s + W_{h^s r^s} h_{t-1}^s + b_{r^s}^s)$$
$$r_t^t = \sigma(W_{x^t r^t} x_t^t + W_{h^t r^t} h_{t-1}^t + b_{r^t}^t)$$
$$u_t^s = \sigma(W_{x^s u^s} x_t^s + W_{h^s u^s} h_{t-1}^s + b_{u^s}^s)$$
$$u_t^t = \sigma(W_{x^t u^t} x_t^t + W_{h^t u^t} h_{t-1}^t + b_{u^t}^t)$$
$$\tilde{h}_t^s = \tan h(W_{x^s h^s} x_t^s + W_{h^s h^s}(r_t^s \odot h_{t-1}^s) + b_{h^s}^s) \quad (14)$$
$$\tilde{h}_t^t = \tan h(W_{x^t h^t} x_t^t + W_{h^t h^t}(r_t^t \odot h_{t-1}^t) + b_{h^t}^t)$$
$$h_t^s = u_t^s \odot \tilde{h}_t^s + (1 - u_t^s) \odot h_{t-1}^s$$
$$h_t^t = u_t^t \odot \tilde{h}_t^t + (1 - u_t^t) \odot h_{t-1}^t$$

## 3.3 Correlation Recurrent Unit (CRU)

### 3.3.1 Autocorrelation Gate

Autocorrelation gate estimates the autocorrelation between a new input $x_t$ and hidden state $h_{t-1}$ within the same time-series components. There are two types of autocorrelation gates, $\hat{a}_t^s$ for seasonality and $\hat{a}_t^t$ for trend. $\hat{a}_t^s$ estimates the autocorrelation between the $x_t^s$ and $h_{t-1}^s$ and $\hat{a}_t^t$ estimates the autocorrelation between the $x_t^t$ and $h_{t-1}^t$. These two gates operate as shown in Eq. (15).

$$\hat{a}_t^s = \sigma(W_{x^s h^s} x_t^s + W_{h^s h^s} h_{t-1}^s + b_{\hat{a}^s}^s)$$
$$\hat{a}_t^t = \sigma(W_{x^t h^t} x_t^t + W_{h^t h^t} h_{t-1}^t + b_{\hat{a}^t}^t) \quad (15)$$

### 3.3.2 Correlation Gate

A correlation gate is used to update the correlation between a new input $x_t$ and hidden state $h_{t-1}$ between different time-series components. $\hat{c}_t^s$ estimates the correlation between the $x_t^s$ and $h_{t-1}^t$ and $c_t^t$ estimates the correlation between $x_t^t$ and $h_{t-1}^s$.

$$\hat{c}_t^s = \sigma(W_{x^s h^s} x_t^s + W_{h^s h^s} h_{t-1}^t + b_{\hat{c}^s}^s)$$
$$\hat{c}_t^s = \sigma(W_{x^s h^s} x_t^s + W_{h^s h^s} h_{t-1}^t + b_{\hat{c}^s}^s) \quad (16)$$
$$\hat{c}_t^t = \sigma(W_{x^t h^t} x_t^t + W_{h^t h^t} h_{t-1}^s + b_{\hat{c}^t}^t)$$

$\hat{c}_t^s$ and $\hat{c}_t^t$ are updated using Eq. (16). All autocorrelation and correlation gates are output through the sigmoid function and have values between 0 and 1. Finally, the values obtained through these gates are used to update the entire hidden state.

### 3.3.3 Correlation Recurrent Gate

The hidden state of the CRU consists of three-dimensional hidden states, $h^t$, $h^s$ and $h^r$. Among them, $h^t$ and $h^s$ are updated recursively; however, $h^r$ is not updated recursively and is updated using the input at point $t$. The hidden state $h_t^r$ for the $x_t^r$ updates the $h_t^r$ using only the remaining component for the $x_t$. The other types of hidden states ($h_t^s$ and $h_t^t$) are updated using Eq. (17) and (18).

$$h_t^s = \lambda \times \tan h(W_{x^s h^s} x_t^s + W_{h^s}(\hat{a}_t^s \odot h_{t-1}^s) + b_{h^s}^s)$$
$$+(1-\lambda) \times \tan h(W_{x^s h^s} x_t^s + W_{h^t h^t}(\hat{c}_t^s \odot h_{t-1}^t) + b_{h^s}^s) \quad (17)$$
$$h_t^t = \lambda \times \tan h(W_{x^t h^t} x_t^t + W_{h^t}(\hat{a}_t^t \odot h_{t-1}^t) + b_{h^t}^t)$$
$$+(1-\lambda) \times \tan h(W_{x^t h^t} x_t^t + W_{h^s h^s}(\hat{c}_t^t \odot h_{t-1}^s) + b_{h^t}^t) \quad (18)$$

Here, each $h^t$ and $h^s$ consists of a weighted sum of the autocorrelation within the component and correlation with the other component. $h_t^s$ combines the effect of autocorrelation in seasonality at time $t$ and the effect of correlation between seasonality and trend. In Eq. (17), if $\hat{a}_t^s$ has a value close to 0, $\hat{a}_t^s \odot h_{t-1}^s$ also has a value close to 0. Therefore, if the autocorrelation between $x_t^s$ and $h_{t-1}^s$ is low, $x_t^s$ is given a high weight to update the $h_t^s$. The remainder of the first term in Eq. (17) operates similarly to the description above. If the correlation between $x_t^s$ and $h_{t-1}^t$ is low, the value of $c_t^s$ is close to zero, and certainly, $\hat{a}_t^s \odot h_{t-1}^t$ also has a value close to zero. Likewise, if the correlation between $x_t^s$ and $h_{t-1}^t$ is low, $x_t^s$ is given a high weight to update $h_t^s$. Finally, $h_t^s$ updates the weighted sum of the left and right terms of the first equation of Eq. (17). $\lambda$ of Eq. (17) is a parameter that determines the ratio of the weighted sum of the two terms and has a value between 0 and 1. If $\lambda$ is updated using 0.5, the two terms are updated with the same weighted sum.

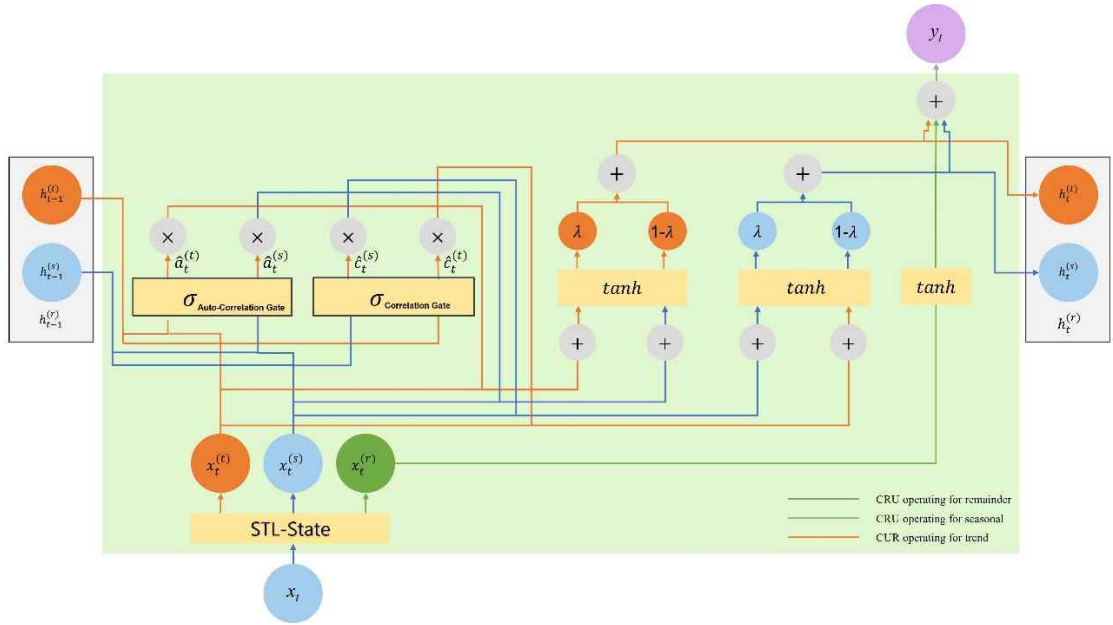

Fig. 6. Correlation Recurrent Unit Cell Structure

In summary, the proposed method has the specificity of a learning mechanism that considers autocorrelation within time-series components and relationships between different components when updating the hidden state. Fig. 6 shows the architecture of CRU.

## 4 EXPERIMENTS

In this section, a comparative experiment is conducted to investigate the superiority of the proposed method. We used five univariate time-series datasets and four multivariate time-series datasets for the experiments. The time-series data used in the experiment are summarized in Fig.7 and 8, and detailed descriptions of the data are provided in Table 1.

We conduct the following four experiments using the nine datasets.

- EXP-I: Comparison of predictive performance in univariate time-series dataset.
- EXP-II: Comparison of predictive performance in multivariate time-series dataset.
- EXP-III: Comparison of parameter size in learning networks.
- EXP-IV: Comparison of predictive performance between CRU and attention mechanisms.

As described in Subsection 3.1, we made STL decomposition operate within the cell structure in a neural net-work. We developed STL cell (STLC) models for the existing RNN, LSTM, and GRU. Thus, we created a new RNN structure with STL cell (RNN-STLC), LSTM with STL cell (LSTM-STLC), and GRU with STL cell (GRU-STLC). We also developed a new neural structure, the CRU. In the experiments, we compared the performance of the CRU with that of existing methods. We used six existing network structures for comparison. 1) RNN, 2) LSTM, 3) GRU, 4) RNNs with STL decomposition (RNN-STLD), 5) LSTMs with STL decomposition (LSTM-STLD), and 6) GRUs with STL decomposition (GRU-STLD). Here, STL decomposition model (STLD) means a network in which each component was learned after decomposing into time-series components using STL. For example, the RNN-STLD shown in Fig. 1 is a network that has decomposed input components.

From EXP-I to EXP-III, we compared the performance of the proposed models (RNN-STLC, LSTM-STLC, GRU-LSTMC, and CRU) with six existing models (RNN, LSTM, GRU, RNN-STLD, LSTM-STLD, and GRU-STLD). All networks used in the experiments used only a single layer with 512 hidden dimensions. Recently, several studies have proven that the attention-based deep learning model shows good performance in multivariate time-series datasets. In this study, a comparative experiment (EXP-IV) was conducted with attention-based deep learning models to prove the superiority of CRU. EXP-IV compares LSTNet-A (long-short time-series network with attention) [37] and DSANet (dual self-attention network) [38] as baseline models with the proposed models. Table 2 lists the models used in each experiment and their characteristics.

## • TABLE 1
### DESCRIPTIONS OF DATASETS FOR THE EXPERIMENTS

| Data Type | Notation | Num. Rows | Num. Columns | Collected Duration | Unit | Data Descript |
|---|---|---|---|---|---|---|
| Unvariate | UTD1 | 294 | 1 | 1947.01.01~ 2020.01.01 | Quarterly | The gross domestic product (GDP) of the United State provided by World Bank Open Data |
| | UTD2 | 2268 | 1 | 2011.11.03~ 2021.12.02 | Daily | S&P500 index is a representative stock index and was collected from Yahoo! Finance Data. |
| | UTD3 | 917 | 1 | 2014.03.27~ 2017.11.10 | Daily | As Google's stock price data, it was collected from Yahoo! Finance Data. |
| | UTD4 | 4567 | 1 | 1999.11.01~ 2018.02.09 | Daily | The Baltic Dry Index (BDI) as major freight indices for dry cargo provides by Batic Exchange |
| | UTD5 | 3298 | 1 | 2008.07.04~ 2021.10.20 | Daily | EU Emssions Trading System (EU-ETS) price provides by EU Transaction Log |
| Multivariate | MTD1 | 2268 | 7 | 2012.12.03~ 2021.12.02 | Daily | Stock prices of seven major US big-tech companies provides by Yahoo! Finace Data. |
| | MTD2 | 3151 | 4 | 2005.07.01~ 2018.02.09 | Daily | BDI, BCI, BPI, and BSI are included as freight indices for dry cargo provided by the Baltic Exchange. |
| | MTD3 | 2140 | 11 | 2011.11.04~ 2021.12.02 | Daily | The eleven-market global stock index (Stock) was collected from Yahoo! Finace Data |
| | MTD4 | 373 | 6 | 1988.01.01~ 2018.12.31 | Monthly | Export and import volumes to six major countries provides by the World Trade Organization (WTO). |

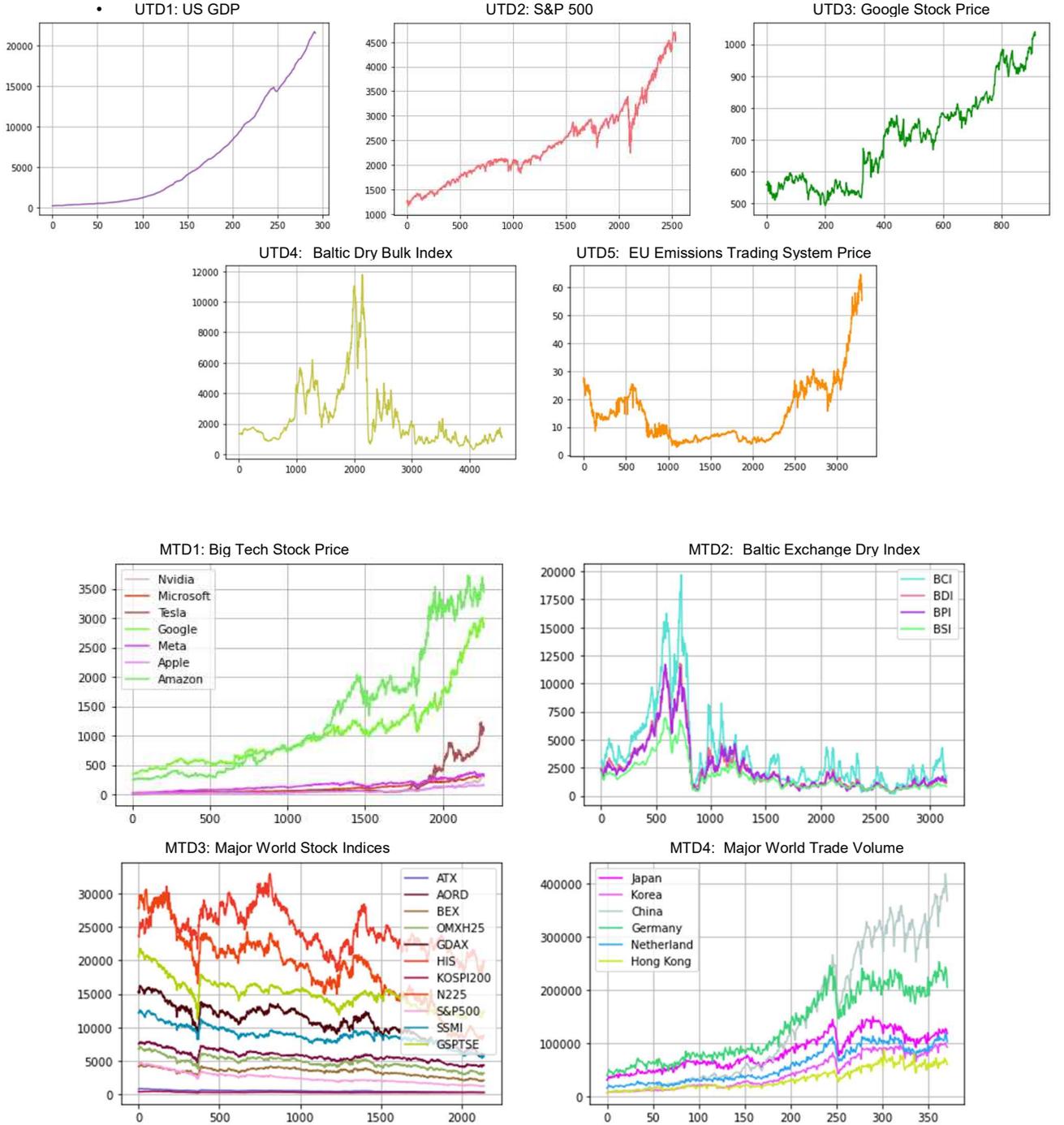

To evaluate the performance of EXP-I, EXP-II, and EXP-IV, we used the root mean squared error (RMSE) and mean absolute percentage error (MAPE). The RMSE and MAPE are given by Eq. (18).

$$\text{RMSE} = \sqrt{\frac{\sum_{n=1}^{N} \sum_{t=1}^{T} (Y_{n,t} - \hat{Y}_{n,t})^2}{N \times T}}$$

$$\text{MAPE} = \sqrt{\frac{\sum_{n=1}^{N} \sum_{t=1}^{T} |(Y_{n,t} - \hat{Y}_{n,t})/Y_{n,t} \times 100|}{N \times T}}$$

(18)

where $Y_{n,t}$ and $\hat{Y}_{n,t}$ are the ground-truth and model-prediction values at time $t$ in the $n^{\text{th}}$ column, respectively. For RMSE and MAPE, a lower value was better.

The ratio of the training and test datasets was split into 70% to 30% and training and validation were performed. All experiments were repeated 15 times. Thousand repetitions of learning were performed under the same conditions. The RMSE and MAPE were measured using the test datasets in the learned model. The errors of the test dataset were measured for each learning and the best values were used for comparison.

The experiments were conducted using a standalone

AMD Ryzen 9 3950X 16-core (3.49 GHz) computer with 128 GB of memory running on Windows 10. The model training was performed using the NVIDIA GeForce TITAN RTX with 24 GB of dedicated GPU memory using CUDA v.11.1 and PyTorch 1.9.0

## 4.1 Comparative Experiments using Univariate Time-series Data

Tables 3 to 6 show the results of EXP-I, and the values in the table show the mean and standard deviation of the RMSE and MAPE of the test data obtained through repeated experiments, respectively (the values in parentheses indicate the standard deviation).

Table 3 shows the results of the short-term predictive performance. Numerous studies have demonstrated that

RNNs which combine STL decomposition methods show improved predictive performance compared to general RNNs [25]-[27]. In the experiment, the application of the STL decomposition method in LSTM and GRU showed improved predictive performance. The proposed STLC models showed improved predictive performance in all experiments compared to when the STLD method was applied. From these experimental results, it is evident that learning decomposed time-series components at the cell level is more effective than learning at the network level. The CRU model, which introduced autocorrelation and correlation gates, showed the best predictive performance. CRU has a 20% lower RMSE and 25% lower MAPE compared to the original RNN, LSTM, and GRU. Additionally, CRU shows approximately 10% better predictive performance than the

• TABLE 2
CONFIGURATION OF MODELS FOR EXPERIENTS

| Model | | Model Characteristics | Experients | | | |
|---|---|---|---|---|---|---|
| | | | EXP-I | EXP-II | EXP-III | EXP-IV |
| **CRU*** | | STL Cell + Autocorrelation Gate + Correlation Gate | ✓ | ✓ | ✓ | ✓ |
| **With STL Cell (STLC models)** | ***RNN-STLC** | RNN + STL Cell | ✓ | ✓ | ✓ | |
| | ***LSTM-STLC** | LSTM + STL Cell | ✓ | ✓ | ✓ | |
| | ***GRU-STLC** | GRU+ STL Cell | ✓ | ✓ | ✓ | |
| With STL Decomposition (STLD models) | RNN-STLD | Multiple RNNs + STL Decomposition Method | ✓ | ✓ | ✓ | |
| | LSTM-STLD | Multiple LSMTs + STL Decomposition Method | ✓ | ✓ | ✓ | |
| | GRU-STLD | Multiple GRUs + STL Decomposition Method | ✓ | ✓ | ✓ | |
| Without STL Decomposition (Original models) | RNN | Original RNN | ✓ | ✓ | ✓ | |
| | LSTM | Orignal LSTM | ✓ | ✓ | ✓ | |
| | GRU | Original GRU | ✓ | ✓ | ✓ | |
| With Attention Mechanism | LSTNet-A | Convlution + RNN + AR + Self-Attention | | | | ✓ |
| | DSANet | Global /Local Temporal Covolution + AR + Self-Attention | | | | ✓ |

*Proposed method

TABLE 3
EXP-I RESULTS OF ONE-STEP-AHEAD (SHORT-TERM) PREDICTION OF UNIVARIATE TIME-SERIES DATA

| Model | UTD1 | | UTD2 | | UTD3 | | UTD4 | | UTD5 | |
|---|---|---|---|---|---|---|---|---|---|---|
| | RMSE | MAPE | RMSE | MAPE | RMSE | MAPE | RMSE | MAPE | RMSE | MAPE |
| **CRU*** | **104.766** | **0.491** | **9.889** | **0.456** | **7.296** | **0.615** | **20.093** | **1.595** | **0.676** | **1.841** |
| | **(1.904)** | **(0.012)** | **(0.010)** | **(0.001)** | **(0.001)** | **(0.001)** | **(0.030)** | **(0.003)** | **(0.006)** | **(0.022)** |
| RNN-STLC | 110.712 | 0.543 | 10.420 | 0.506 | 7.678 | 0.665 | 25.409 | 1.673 | 0.714 | 1.898 |
| | (2.425) | (0.014) | (0.013) | (0.001) | (0.001) | (0.001) | (0.041) | (0.004) | (0.004) | (0.016) |
| LSTM-STLC | 109.973 | 0.539 | 10.790 | 0.521 | 7.730 | 0.673 | 26.298 | 1.746 | 0.712 | 1.890 |
| | (2.832) | (0.016) | (0.108) | (0.006) | (0.002) | (0.001) | (0.224) | (0.018) | (0.007) | (0.022) |
| GRU-STLC | 110.752 | 0.545 | 10.503 | 0.508 | 7.680 | 0.665 | 25.491 | 1.681 | 0.710 | 1.889 |
| | (2.081) | (0.013) | (0.043) | (0.002) | (0.001) | (0.001) | (0.076) | (0.007) | (0.005) | (0.017) |
| RNN-STLD | 112.889 | 0.539 | 11.337 | 0.645 | 8.228 | 0.706 | 26.537 | 1.720 | 0.796 | 2.074 |
| | (0.118) | (0.012) | (0.789) | (0.031) | (0.002) | (0.001) | (0.006) | (0.007) | (0.001) | (0.006) |
| LSTM-STLD | 115.309 | 0.552 | 12.018 | 0.687 | 8.230 | 0.707 | 26.539 | 1.716 | 0.808 | 2.105 |
| | (0.457) | (0.008) | (1.494) | (0.021) | (0.001) | (0.001) | (0.001) | (0.001) | (0.001) | (0.006) |
| GRU-STLD | 113.888 | 0.531 | 11.304 | 0.641 | 8.230 | 0.705 | 26.539 | 1.716 | 0.808 | 2.104 |
| | (0.533) | (0.005) | (0.652) | (0.010) | (0.001) | (0.001) | (0.001) | (0.001) | (0.001) | (0.011) |
| RNN | 120.707 | 0.872 | 12.283 | 0.579 | 9.016 | 0.734 | 30.689 | 2.110 | 0.863 | 2.279 |
| | (3.183) | (0.124) | (0.012) | (0.001) | (0.001) | (0.001) | (0.016) | (0.007) | (0.004) | (0.021) |
| LSTM | 121.462 | 0.876 | 12.333 | 0.583 | 9.046 | 0.736 | 30.623 | 1.984 | 0.868 | 2.291 |
| | (3.476) | (0.012) | (0.022) | (0.002) | (0.001) | (0.001) | (0.008) | (0.005) | (0.001) | (0.001) |
| GRU | 121.437 | 0.775 | 12.363 | 0.584 | 9.048 | 0.736 | 30.570 | 2.079 | 0.862 | 2.272 |
| | (3.455) | (0.101) | (0.015) | (0.001) | (0.001) | (0.001) | (0.006) | (0.007) | (0.006) | (0.032 |

Avg. (Std. Dev.)

TABLE 4
EXP-I RESULTS OF THREE-STEP-AHEAD PREDICTION OF UNIVARIATE TIME-SERIES DATA

| Model | UTD1 | | UTD2 | | UTD3 | | UTD4 | | UTD5 | |
|---|---|---|---|---|---|---|---|---|---|---|
| | RMSE | MAPE | RMSE | MAPE | RMSE | MAPE | RMSE | MAPE | RMSE | MAPE |
| **CRU\*** | **180.896** (0.736) | **0.943** (0.002) | **14.671** (0.003) | **0.693** (0.001) | **11.331** (0.001) | **1.002** (0.001) | **54.862** (0.021) | **3.952** (0.002) | **0.939** (0.001) | **2.657** (0.006) |
| RNN-STLC | 190.632 (0.859) | 0.994 (0.003) | 15.455 (0.012) | 0.744 (0.001) | 11.927 (0.001) | 1.052 (0.001) | 57.788 (0.031) | 4.002 (0.002) | 0.988 (0.001) | 2.719 (0.005) |
| LSTM-STLC | 190.707 (0.787) | 0.995 (0.003) | 16.175 (0.159) | 0.795 (0.012) | 11.927 (0.002) | 1.052 (0.001) | 58.078 (0.122) | 4.012 (0.006) | 1.004 (0.003) | 2.724 (0.007) |
| GRU-STLC | 190.516 (0.442) | 0.993 (0.002) | 15.939 (0.141) | 0.781 (0.010) | 11.928 (0.002) | 1.051 (0.001) | 57.881 (0.030) | 4.005 (0.003) | 0.999 (0.003) | 2.712 (0.006) |
| RNN-STLD | 203.119 (2.728) | 1.163 (0.038) | 17.255 (0.967) | 0.925 (0.053) | 14.258 (0.003) | 1.255 (0.001) | 67.539 (0.027) | 4.648 (0.069) | 1.184 (0.001) | 3.292 (0.012) |
| LSTM-STLD | 208.232 (2.252) | 1.128 (0.015) | 17.646 (1.335) | 0.952 (0.075) | 14.245 (0.001) | 1.255 (0.001) | 67.677 (0.001) | 4.646 (0.001) | 1.183 (0.002) | 3.239 (0.005) |
| GRU-STLD | 209.512 (1.837) | 1.145 (0.052) | 18.528 (1.300) | 1.008 (0.073) | 14.244 (0.001) | 1.255 (0.001) | 67.433 (0.020) | 4.643 (0.023) | 1.184 (0.002) | 3.239 (0.004) |
| RNN | 221.559 (1.804) | 1.087 (0.034) | 20.742 (0.042) | 1.012 (0.001) | 16.065 (0.001) | 1.402 (0.001) | 69.428 (0.052) | 4.862 (0.068) | 1.356 (0.001) | 3.636 (0.010) |
| LSTM | 223.412 (1.909) | 1.039 (0.033) | 20.806 (0.076) | 1.014 (0.003) | 16.247 (0.001) | 1.428 (0.001) | 69.400 (0.049) | 4.846 (0.033) | 1.388 (0.001) | 3.738 (0.014) |
| GRU | 223.322 (1.144) | 1.049 (0.031) | 20.941 (0.039) | 1.022 (0.001) | 16.217 (0.009) | 1.423 (0.001) | 69.388 (0.038) | 4.843 (0.034) | 1.377 (0.002) | 3.680 (0.008) |

Avg. (Std. Dev.)

TABLE 5
EXP-I RESULTS OF SIX-STEP-AHEAD PREDICTION OF UNIVARIATE TIME-SERIES DATA

| Model | UTD1 | | UTD2 | | UTD3 | | UTD4 | | UTD5 | |
|---|---|---|---|---|---|---|---|---|---|---|
| | RMSE | MAPE | RMSE | MAPE | RMSE | MAPE | RMSE | MAPE | RMSE | MAPE |
| **CRU\*** | **307.031** (1.529) | **1.607** (0.003) | **21.661** (0.024) | **1.031** (0.001) | **16.776** (0.008) | **1.506** (0.001) | **93.578** (0.024) | **6.745** (0.004) | **1.515** (0.009) | **4.320** (0.024) |
| RNN-STLC | 322.892 (1.886) | 1.756 (0.005) | 22.853 (0.034) | 1.084 (0.002) | 17.638 (0.005) | 1.558 (0.001) | 98.594 (0.086) | 6.803 (0.007) | 1.572 (0.008) | 4.326 (0.007) |
| LSTM-STLC | 323.358 (1.749) | 1.757 (0.007) | 26.132 (0.281) | 1.312 (0.019) | 18.648 (0.027) | 1.676 (0.001) | 101.859 (0.172) | 7.029 (0.018) | 1.642 (0.006) | 4.529 (0.014) |
| GRU-STLC | 323.240 (1.550) | 1.757 (0.006) | 25.205 (0.209) | 1.258 (0.015) | 17.747 (0.020) | 1.560 (0.001) | 98.673 (0.094) | 6.817 (0.015) | 1.629 (0.005) | 4.481 (0.023) |
| RNN-STLD | 344.132 (1.566) | 2.056 (0.041) | 24.792 (0.514) | 1.310 (0.010) | 21.390 (0.006) | 1.900 (0.001) | 113.888 (0.406) | 7.974 (0.119) | 1.853 (0.020) | 5.315 (0.108) |
| LSTM-STLD | 316.458 (3.314) | 1.847 (0.006) | 26.108 (0.356) | 1.421 (0.093) | 20.818 (0.005) | 1.896 (0.001) | 113.946 (0.508) | 8.145 (0.212) | 1.803 (0.016) | 5.026 (0.115) |
| GRU-STLD | 339.515 (6.784) | 1.900 (0.004) | 25.834 (0.415) | 1.402 (0.064) | 20.821 (0.003) | 1.897 (0.001) | 113.901 (0.032) | 7.943 (0.022) | 1.812 (0.018) | 5.172 (0.103) |
| RNN | 366.483 (2.136) | 1.892 (0.031) | 28.783 (0.019) | 1.407 (0.001) | 22.616 (0.041) | 1.980 (0.004) | 118.033 (0.066) | 8.155 (0.029) | 1.845 (0.005) | 5.192 (0.027) |
| LSTM | 358.729 (1.992) | 1.798 (0.034) | 28.906 (0.244) | 1.417 (0.027) | 23.017 (0.001) | 2.008 (0.001) | 118.159 (0.086) | 8.172 (0.033) | 1.906 (0.003) | 5.400 (0.019) |
| GRU | 374.233 (2.052) | 1.949 (0.041) | 29.551 (0.172) | 1.484 (0.016) | 22.762 (0.015) | 1.986 (0.001) | 118.131 (0.059) | 8.149 (0.019) | 1.860 (0.007) | 5.243 (0.022) |

Avg. (Std. Dev.)

STLD models of RNN-STLD, LSTM-STLD, and GRU-STLD. The experimental results show that it is effective to learn by introducing the autocorrelation and correlation gates while considering the components of time-series decomposition.

## 4.2 Comparative Experiment using Multivariate Time-series Data

Tables 7–10 show the results of EXP-II using multivariate time-series data. In the results of EXP-II, CRU showed the best RMSE and MAPE values for all multivariate time-series data. We discovered that the performance improvement of the proposed model in multivariate time-series data is larger than that in the univariate model.

Table 7 shows that the RMSE and MAPE of short-term prediction by the CRU model showed an improvement of 5% to 30% compared to other networks. In terms of long-term prediction, as presented in Tables 8–10, CRU showed an average of 20% improvement in RMSE and MAPE compared to other models.

TABLE 6
EXP-I RESULTS OF TWELVE-STEP-AHEAD (SHORT-TERM) PREDICTION OF UNIVARIATE TIME-SERIES DATA

| Model | UTD1 | | UTD2 | | UTD3 | | UTD4 | | UTD5 | |
|---|---|---|---|---|---|---|---|---|---|---|
| | RMSE | MAPE | RMSE | MAPE | RMSE | MAPE | RMSE | MAPE | RMSE | MAPE |
| **CRU*** | **498.986** | **2.816** | **35.493** | **1.780** | **27.065** | **2.455** | **160.239** | **11.397** | **2.375** | **6.868** |
| | **(2.067)** | **(0.004)** | **(0.193)** | **(0.014)** | **(0.036)** | **(0.002)** | **(0.056)** | **(0.042)** | **(0.041)** | **(0.030)** |
| RNN-STLC | 524.223 | 2.865 | 36.219 | 1.829 | 28.351 | 2.497 | 175.063 | 12.431 | 2.454 | 6.771 |
| | (3.191) | (0.005) | (0.277) | (0.021) | (0.029) | (0.002) | (0.070) | (0.029) | (0.053) | (0.040) |
| LSTM-STLC | 525.073 | 2.867 | 44.019 | 2.340 | 29.042 | 2.541 | 177.974 | 12.954 | 2.582 | 7.205 |
| | (3.306) | (0.007) | (0.305) | (0.021) | (0.030) | (0.002) | (0.369) | (0.050) | (0.010) | (0.036) |
| GRU-STLC | 524.540 | 2.863 | 43.534 | 2.312 | 28.841 | 2.526 | 175.921 | 12.714 | 2.578 | 7.189 |
| | (2.596) | (0.005) | (0.284) | (0.020) | (0.035) | (0.002) | (0.141) | (0.048) | (0.006) | (0.018) |
| RNN-STLD | 528.866 | 3.167 | 37.584 | 2.041 | 29.483 | 2.659 | 185.801 | 13.135 | 2.568 | 7.569 |
| | (5.199) | (0.008) | (0.210) | (0.026) | (0.010) | (0.001) | (0.243) | (0.178) | (0.045) | (0.067) |
| LSTM-STLD | 532.434 | 3.001 | 37.200 | 2.019 | 28.887 | 2.591 | 184.573 | 13.223 | 2.558 | 7.692 |
| | (3.678) | (0.020) | (2.019) | (0.034) | (0.008) | (0.001) | (0.275) | (0.180) | (0.044) | (0.030) |
| GRU-STLD | 538.548 | 3.067 | 39.261 | 2.185 | 29.089 | 2.579 | 185.468 | 13.159 | 2.542 | 7.606 |
| | (3.847) | (0.019) | (2.185) | (0.090) | (0.006) | (0.001) | (1.087) | (0.060) | (0.025) | (0.041) |
| RNN | 557.918 | 2.955 | 38.659 | 1.907 | 30.890 | 2.676 | 210.236 | 14.948 | 2.750 | 8.059 |
| | (4.167) | (0.060) | (0.012) | (0.004) | (0.080) | (0.018) | (0.058) | (0.056) | (0.023) | (0.159) |
| LSTM | 541.198 | 2.763 | 39.419 | 1.910 | 30.762 | 2.712 | 210.177 | 15.019 | 2.824 | 8.531 |
| | (4.061) | (0.042) | (0.306) | (0.016) | (0.045) | (0.005) | (0.303) | (0.138) | (0.020) | (0.040) |
| GRU | 528.484 | 2.723 | 44.963 | 2.383 | 30.824 | 2.698 | 210.650 | 15.051 | 2.781 | 8.398 |
| | (4.109) | (0.066) | (1.137) | (0.073) | (0.029) | (0.016) | (0.605) | (0.125) | (0.016) | (0.033) |

Avg. (Std. Dev.)

EXP-I and EXP-II show that the proposed STLC models and CRU demonstrate improved predictive performance over six networks in all period predictions (1-step-ahead, 3-step-ahead, 6-step-ahead, and 12-step-ahead) regardless of univariate or multivariate time-series prediction.

### 4.3 Comparison of Parameter Size

Although the proposed model outperforms the existing models, the number of parameters increases because more parameters need to be estimated in processing each decomposition component. Therefore, this subsection evaluates the parameter sizes of the proposed models. We compared the parameter sizes of the proposed models: RNN-STLC, LSTM-STLC, and GRU-STLC.

Fig. 9 shows the comparison results of the parameters of the entire network used in EXP-I and EXP-II. When STLD is used, the number of parameters increases by approximately two times that of the original RNN, LSTM, and GRU models. Despite the increase in the number of parameters, STLD has been widely used because it can achieve better performance.

In this regard, we compared the parameter sizes of STLC and STLD. Fig. 9 shows that STLC has a 30% smaller number of parameters and better performance compared to STLD. The CRU model uses a smaller number of parameters than the STLD models except for RNN-STLC. Notably, CRU has 21% and 41% fewer parameters than LSTM-STLD and GRU-STLD, respectively. In summary, the CRU and STLC models proposed in this study use a smaller number of parameters and show better performance than the existing STLD models.

### 4.4 Comparison with Attention Mechanism

In this subsection, we present comparative experiments (EXP-IVs) conducted using two attention-based deep learning models (LSTNet-A and DSANet), which have been widely used to improve the predictive accuracy of TSF problems [39]. These two networks show good results in TSF problems using multivariate time-series data and are used as baseline models in various studies [40]-[42].

Tables 11–14 show the comparison results. CRU showed improved RMSE and MAPE in all multivariate time-series data in the short-and long-term predictions compared to LSTNet. Compared to DSANet, CRU showed improved predictive performance in all three predictive scenarios. In the case of a 12-step-ahead prediction, the proposed model still showed better performance in MTD2 and MTD4. However, in the MTD1 and MTD3 datasets, DSANet showed better performance.

In time-series data, the attention mechanism is a method for capturing related patterns that have occurred in the past and using them to learn current time-series data. This approach increases predictive performance

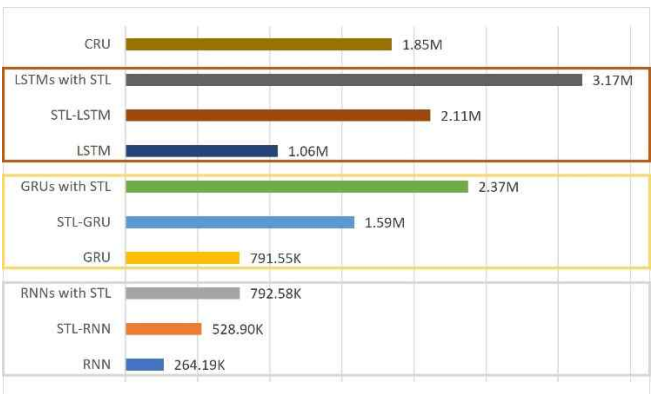

Fig. 9. Average Number of Parameters of the Trained Model used as EXP-I and EXP-II

## TABLE 7
### EXP-II Results of One-Step-Ahead (Short-Term) Prediction of Multivariate Time-Series Data

| | MTD1 | | MTD2 | | MTD3 | | MTD4 | |
|---|---|---|---|---|---|---|---|---|
| | RMSE | MAPE | RMSE | MAPE | RMSE | MAPE | RMSE | MAPE |
| **CRU\*** | **20.15 (0.41)** | **1.56 (0.03)** | **42.63 (0.16)** | **2.00 (0.04)** | **95.17 (3.17)** | **0.91 (0.04)** | **9556.8 (455.7)** | **5.22 (0.18)** |
| RNN-STLC | 21.28 (0.80) | 1.64 (0.06) | 47.51 (0.22) | 2.58 (0.05) | 107.55 (3.74) | 1.02 (0.05) | 10462.1 (579.1) | 5.37 (0.26) |
| LSTM-STLC | 24.63 (1.70) | 1.71 (0.31) | 46.30 (0.17) | 2.31 (0.12) | 115.91 (5.82) | 1.14 (0.08) | 10559.3 (354.6) | 5.36 (0.20) |
| GRU-STLC | 22.21 (1.52) | 1.69 (0.05) | 48.00 (0.43) | 2.69 (0.11) | 111.80 (5.61) | 1.09 (0.11) | 10606.5 (431.7) | 5.40 (0.18) |
| RNN-STLD | 25.38 (2.45) | 1.94 (0.12) | 47.75 (0.33) | 2.61 (0.41) | 190.65 (6.88) | 1.95 (0.15) | 11115.8 (541.9) | 6.13 (0.56) |
| LSTM-STLD | 24.91 (1.98) | 1.82 (0.09) | 47.05 (0.15) | 2.45 (0.18) | 184.89 (5.98) | 1.81 (0.08) | 10905.1 (336.8) | 5.85 (0.45) |
| GRU-STLD | 25.01 (2.22) | 1.88 (0.10) | 47.69 (0.21) | 2.54 (0.25) | 181.74 (5.51) | 1.78 (0.12) | 10987.6 (401.2) | 5.77 (0.47) |
| RNN | 29.66 (2.87) | 3.11 (0.39) | 48.65 (0.02) | 2.70 (0.03) | 212.12 (8.72) | 2.30 (0.22) | 12701.9 (468.4) | 8.80 (0.50) |
| LSTM | 26.85 (1.45) | 2.42 (0.08) | 48.05 (0.03) | 2.65 (0.02) | 202.98 (6.54) | 2.01 (0.19) | 12696.7 (462.3) | 7.99 (0.28) |
| GRU | 26.37 (1.83) | 2.51 (0.27) | 48.19 (0.02) | 2.75 (0.02) | 205.02 (9.67) | 2.16 (0.15) | 12719.2 (444.9) | 8.05 (0.13) |

Avg. (Std. Dev.)

## TABLE 8
### EXP-II Results of Three-Step-Ahead Prediction of Multivariate Time-Series Data

| | MTD1 | | MTD2 | | MTD3 | | MTD4 | |
|---|---|---|---|---|---|---|---|---|
| | RMSE | MAPE | RMSE | MAPE | RMSE | MAPE | RMSE | MAPE |
| **CRU\*** | **34.04 (1.86)** | **2.75 (0.07)** | **97.02 (0.36)** | **6.49 (0.13)** | **158.07 (1.93)** | **1.61 (0.07)** | **14018.8 (520.0)** | **8.04 (0.21)** |
| RNN-STLC | 36.39 (3.01) | 2.84 (0.10) | 107.56 (0.50) | 6.93 (0.18) | 171.17 (3.84) | 1.72 (0.10) | 15670.1 (762.5) | 8.13 (0.43) |
| LSTM-STLC | 45.66 (3.58) | 3.22 (0.11) | 112.56 (0.93) | 7.05 (0.21) | 173.78 (3.31) | 1.73 (0.07) | 15681.1 (627.4) | 8.14 (0.47) |
| GRU-STLC | 38.24 (2.72) | 2.85 (0.07) | 108.54 (0.47) | 7.25 (0.17) | 175.47 (4.37) | 1.74 (0.14) | 16032.4 (587.1) | 8.39 (0.33) |
| RNN-STLD | 48.34 (3.21) | 3.75 (0.22) | 115.68 (0.58) | 7.26 (0.22) | 320.23 (7.65) | 3.09 (0.37) | 20031.9 (745.1) | 11.26 (1.04) |
| LSTM-STLD | 46.93 (3.15) | 3.52 (0.12) | 112.17 (0.49) | 7.15 (0.19) | 298.51 (7.20) | 2.65 (0.15) | 18875.0 (666.6) | 10.47 (0.69) |
| GRU-STLD | 47.10 (2.99) | 3.58 (0.15) | 111.95 (0.37) | 7.12 (0.15) | 307.77 (7.19) | 2.77 (0.20) | 19105.1 (703.5) | 10.95 (0.73) |
| RNN | 58.05 (3.83) | 4.33 (0.13) | 121.67 (0.21) | 7.55 (0.11) | 540.97 (28.8) | 5.06 (0.33) | 26118.7 (864.7) | 13.18 (1.10) |
| LSTM | 51.94 (1.33) | 3.95 (0.10) | 119.24 (0.14) | 7.31 (0.10) | 418.46 (21.4) | 3.82 (0.18) | 25578.6 (660.9) | 12.53 (0.97) |
| GRU | 52.39 (1.10) | 4.72 (0.17) | 119.32 (0.14) | 7.40 (0.05) | 423.65 (23.3) | 4.07 (0.20) | 23691.0 (604.4) | 11.32 (0.74) |

Avg. (Std. Dev.)

## TABLE 9
### EXP-II Results of Six-Step-Ahead Prediction of Multivariate Time-Series Data

| | MTD1 | | MTD2 | | MTD3 | | MTD4 | |
|---|---|---|---|---|---|---|---|---|
| | RMSE | MAPE | RMSE | MAPE | RMSE | MAPE | RMSE | MAPE |
| **CRU\*** | **61.03 (4.66)** | **4.97 (0.21)** | **167.26 (0.69)** | **12.7 (0.28)** | **224.53 (2.33)** | **1.93 (0.08)** | **18321.6 (616.2)** | **9.31 (0.30)** |
| RNN-STLC | 63.77 (6.30) | 5.22 (0.36) | 186.40 (0.83) | 13.8 (0.42) | 234.32 (4.86) | 1.98 (0.10) | 22167.1 (702.1) | 10.66 (0.37) |
| LSTM-STLC | 68.25 (6.25) | 5.40 (0.43) | 192.23 (1.93) | 14.3 (0.39) | 258.49 (4.15) | 2.23 (0.14) | 21287.8 (657.7) | 10.26 (0.40) |
| GRU-STLC | 64.06 (5.74) | 5.11 (0.29) | 189.40 (1.48) | 14.1 (0.40) | 240.40 (3.99) | 2.02 (0.13) | 21598.7 (824.1) | 10.43 (0.28) |
| RNN-STLD | 72.11 (5.74) | 6.47 (0.59) | 193.54 (2.74) | 14.5 (1.33) | 370.05 (6.22) | 4.02 (0.20) | 23579.3 (745.8) | 12.14 (0.52) |
| LSTM-STLD | 70.03 (5.34) | 6.12 (0.46) | 190.10 (2.04) | 14.1 (1.11) | 334.43 (4.84) | 3.69 (0.15) | 23142.6 (698.0) | 11.91 (0.41) |
| GRU-STLD | 69.45 (5.55) | 6.09 (0.30) | 191.95 (2.25) | 14.1 (1.17) | 327.89 (4.30) | 3.57 (0.12) | 23244.5 (721.5) | 12.02 (0.45) |
| RNN | 93.21 (5.21) | 8.05 (0.66) | 195.87 (1.61) | 15.0 (0.45) | 652.70 (25.8) | 7.04 (0.33) | 30793.9 (951.4) | 17.15 (1.53) |
| LSTM | 84.19 (5.55) | 6.91 (0.40) | 195.85 (1.46) | 14.9 (0.40) | 618.04 (28.5) | 6.30 (0.21) | 27641.8 (935.6) | 15.33 (0.90) |
| GRU | 88.35 (5.23) | 7.09 (0.51) | 195.32 (1.14) | 14.9 (0.33) | 636.72 (30.6) | 6.83 (0.18) | 25492.4 (854.7) | 14.65 (0.75) |

Avg. (Std. Dev.)

## TABLE 10
### EXP-II Results of Twelve-Step-Ahead Prediction of Multivariate Time-Series Data

| | MTD1 | | MTD2 | | MTD3 | | MTD4 | |
|---|---|---|---|---|---|---|---|---|
| | RMSE | MAPE | RMSE | MAPE | RMSE | MAPE | RMSE | MAPE |
| **CRU\*** | **96.49 (3.16)** | **8.01 (0.28)** | **270.16 (0.46)** | **18.8 (0.12)** | **499.03 (4.22)** | **4.84 (0.26)** | **20738.9 (755.1)** | **11.69 (0.46)** |
| RNN-STLC | 103.22 (2.51) | 8.50 (0.45) | 290.48 (0.78) | 20.4 (0.14) | 536.64 (7.47) | 5.35 (0.51) | 25738.5 (1020.1) | 13.80 (0.67) |
| LSTM-STLC | 103.07 (7.10) | 8.75 (0.42) | 292.76 (0.83) | 20.4 (0.16) | 577.88 (15.1) | 5.64 (0.57) | 26134.5 (870.5) | 13.72 (0.78) |
| GRU-STLC | 98.76 (7.11) | 8.46 (0.28) | 291.75 (0.74) | 20.3 (0.16) | 553.32 (19.2) | 5.40 (0.60) | 26078.0 (903.3) | 13.72 (0.62) |
| RNN-STLD | 112.45 (8.31) | 9.81 (0.75) | 295.62 (1.01) | 21.2 (0.18) | 670.05 (22.48) | 7.12 (0.84) | 27014.0 (784.6) | 15.00 (0.98) |
| LSTM-STLD | 108.32 (6.29) | 9.05 (0.66) | 294.09 (0.97) | 20.9 (0.15) | 634.17 (19.01) | 6.57 (0.69) | 26858.1 (677.6) | 14.89 (0.77) |
| GRU-STLD | 109.87 (7.87) | 9.12 (0.69) | 293.74 (0.91) | 20.8 (0.14) | 645.98 (19.36) | 6.73 (0.78) | 26971.9 (745.3) | 14.94 (0.84) |
| RNN | 132.11 (9.85) | 14.2 (0.95) | 301.93 (2.54) | 21.9 (0.81) | 922.33 (58.9) | 9.32 (1.02) | 32610.0 (1888.5) | 19.65 (1.25) |
| LSTM | 118.85 (6.68) | 10.3 (0.61) | 299.22 (1.33) | 21.6 (0.93) | 814.05 (40.4) | 8.31 (0.79) | 28092.5 (1294.5) | 16.78 (1.04) |
| GRU | 131.39 (8.82) | 12.4 (0.53) | 301.17 (2.36) | 22.0 (0.86) | 821.95 (51.5) | 8.48 (0.92) | 27475.8 (1092.7) | 15.19 (0.93) |

Avg. (Std. Dev.)

**TABLE 11**
EXP-IV RESULTS OF ONE-STEP-AHEAD (SHORT-TERM) PREDICTION

| | MTD1 | | MTD2 | | MTD3 | | MTD4 | |
|---|---|---|---|---|---|---|---|---|
| | RMSE | MAPE | RMSE | MAPE | RMSE | MAPE | RMSE | MAPE |
| **CRU*** | **20.15 (0.41)** | **1.56 (0.03)** | **42.63 (0.16)** | **2.00 (0.04)** | **95.17 (3.17)** | **0.91 (0.04)** | **9556.8 (455.7)** | **5.22 (0.18)** |
| LSTNet-A | 30.14 (5.35) | 2.99 (0.29) | 47.85 (0.40) | 2.62 (0.10) | 106.98 (5.59) | 1.11 (0.12) | 10256.8 (545.9) | 5.34 (0.27) |
| DSANet | 28.32 (4.93) | 2.78 (0.11) | 47.34 (0.28) | 2.58 (0.07) | 105.34 (5.20) | 1.05 (0.10) | 10159.9 (589.1) | 5.30 (0.20) |

Avg. (Std. Dev.)

**TABLE 12**
EXP-IV RESULTS OF THREE-STEP-AHEAD PREDICTION

| | MTD1 | | MTD2 | | MTD3 | | MTD4 | |
|---|---|---|---|---|---|---|---|---|
| | RMSE | MAPE | RMSE | MAPE | RMSE | MAPE | RMSE | MAPE |
| **CRU*** | **34.04 (1.86)** | **2.75 (0.07)** | **97.02 (0.36)** | **6.49 (0.13)** | **158.07 (1.93)** | **1.61 (0.07)** | **14018.8 (520.0)** | **8.04 (0.21)** |
| LSTNet-A | 54.25 (2.82) | 4.21 (0.45) | 115.72 (0.58) | 7.36 (0.51) | 172.65 (2.49) | 1.89 (0.15) | 14891.8 (587.1) | 8.09 (0.30) |
| DSANet | 53.78 (2.04) | 4.08 (0.36) | 115.18 (0.47) | 7.03 (0.34) | 170.34 (2.33) | 1.80 (0.12) | 14740.0 (601.8) | 8.07 (0.28) |

Avg. (Std. Dev.)

**TABLE 13**
EXP-IV RESULTS OF SIX-STEP-AHEAD PREDICTION

| | MTD1 | | MTD2 | | MTD3 | | MTD4 | |
|---|---|---|---|---|---|---|---|---|
| | RMSE | MAPE | RMSE | MAPE | RMSE | MAPE | RMSE | MAPE |
| **CRU*** | **61.03 (4.66)** | **4.97 (0.21)** | **167.26 (0.69)** | **12.7 (0.28)** | **224.53 (2.33)** | **1.93 (0.08)** | **18321.6 (616.2)** | **9.31 (0.30)** |
| LSTNet-A | 66.02 (3.42) | 5.47 (0.41) | 185.32 (1.37) | 14.9 (0.74) | 247.19 (2.65) | 2.19 (0.25) | 20658.9 (603.5) | 9.45 (0.49) |
| DSANet | 65.82 (3.21) | 5.32 (0.35) | 184.18 (1.18) | 14.2 (0.43) | 245.21 (2.54) | 2.00 (0.15) | 19690.5 (548.4) | 9.41 (0.33) |

Avg. (Std. Dev.)

**TABLE 14**
EXP-IV RESULTS OF TWELVE-STEP-AHEAD PREDICTION

| | MTD1 | | MTD2 | | MTD3 | | MTD4 | |
|---|---|---|---|---|---|---|---|---|
| | RMSE | MAPE | RMSE | MAPE | RMSE | MAPE | RMSE | MAPE |
| **CRU*** | 96.49 (3.16) | 8.01 (0.28) | 270.16 (0.46) | 18.8 (0.12) | 499.03 (4.22) | 4.84 (0.26) | **20738.9 (755.1)** | **11.69 (0.46)** |
| LSTNet-A | 95.98 (2.65) | 7.91 (0.36) | 285.21 (0.91) | 19.5 (0.41) | 440.16 (4.01) | 3.81 (0.29) | 22021.5 (645.2) | 12.01 (0.58) |
| DSANet | **95.27 (2.00)** | **7.88 (0.26)** | 280.56 (0.78) | 19.1 (0.31) | **432.70 (3.76)** | **3.73 (0.28)** | 21316.2 (586.6) | 11.99 (0.36) |

Avg. (Std. Dev.)

when a similar pattern occurs in the learning data. However, the predictive performance is low when large fluctuations that are not included in the learning data occur. Conversely, CRUs ensure a more robust predictive performance because each component (trend, season, and rest) inherent in time-series data is learned in consideration of each other.

## 5 CONCLUSIONS

In this study, we proposed a novel neural architecture called correlation recurrent unit (CRU) to improve the predictive performance of the TSF problem. CRU includes the following new concepts: 1) STL cell: A cell that decomposes the new time-series input data $x_t$ into three time-series components (trend, seasonal, and remainder); 2) autocorrelation and correlation gates: gates that estimate the correlation and autocorrelation existing between the time-series component and hidden layer.

Two factors enable CRU to perform better than other networks: first, the decomposition of the time-series component occurs inside the cell rather than at the network level; therefore, each time-series component that can be learned in detail is inherent in the time-series data; second, a more accurate prediction is possible by learning the correlation and autocorrelation among time-series components. First, we constructed models containing STLC and conducted the experiment; consequently, the first factor was proven. Additionally, by comparing CRUs with existing methods, it was evident that correlation and autocorrelation in CRU structure could contribute to predictive improvement. The CRU model showed a performance improvement of more than 20% compared to the original deep learning model for predicting time series for various data subjects in this study.

In recent years, a model that combines attention mechanisms in multivariate time-series prediction problems has shown good predictive performance. This is similar to the concept of correlation and autocorrelation. Therefore, a comparative experiment was performed using a model that combines the CRU and attention mechanism. The results of the experiment showed that predicting future values using autocorrelation and correlation between time-series components was more effective than using the attention mechanism.

The CRU proposed in this study could be a good alternative to the existing network in TSF issues, and it is expected that CRU will help to make more accurate decisions in areas related to TSF issues.

## ACKNOWLEDGMENT

This research was supported by Basic Science Research

Program through the National Research Foundation of Korea (NRF) funded by the Ministry of Education (NRF-2020R1A2C1102294).

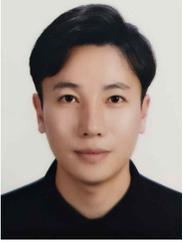

**Sunghyun** Sim received his BS. degree in Statistics from Pusan National University, Rep. of Korea, in 2016, and his MS and PhD degrees in Industrial Engineering in 2021. His research interest process optimization based on the deep-learning method.

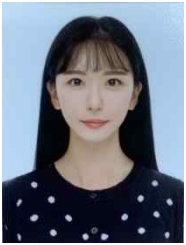

**Dohee Kim** received her BS. degree in Industrial Engineering from Pusan National University, Rep. of Korea, in 2019, and currently, she is working toward the PhD degree at Industrial Engineering, Pusan National University, Rep. of Korea. She is interested in Time series forecasting: Deep learning and Data mining

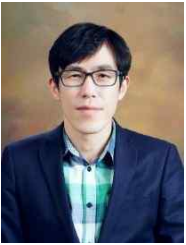

**Hyerim Bae** received his BS, MS, and PhD degrees in Industrial Engineering from Seoul National University, Rep. of Korea. From 2002 to 2003, he worked for Samsung Credit Card Corp., Seoul, Rep. of Korea. Since 2005, he has been a professor with the Department of Industrial Engineering, Pusan National University, Rep. of Korea. He is interested in information system design; cloud computing; business process management systems, and process mining and big data analytics for operational intelligence.His current research activities include analyzing huge volumes of event logs from port logistics and shipbuilding industries using process mining techniques